\documentclass[]{style}
\usepackage{amsmath,amssymb,booktabs,enumitem}
\usepackage[bottom]{footmisc}
\setcitestyle{numbers,square}

\hypersetup{
  colorlinks=true,
  linkcolor=metablue,
  citecolor=metablue,
  urlcolor=metablue
}
\setlist{nosep,leftmargin=*}
\newcommand{\templateicon}[1]{%
  \raisebox{-0.18em}{\includegraphics[height=1.2em]{assets/#1.png}}}

\title{Demo2Skill: Equipping Robo-Use Agents with Dexterous Skills through Demonstration and Simulation}

\author[1,2,\ensuremath{\dagger},*]{Zijian Xiao}
\author[1,\ensuremath{\dagger}]{Jinkun Hao}
\author[3]{Yibin Liu}
\author[2,\ensuremath{\ddagger}]{Zipeng Ye}
\author[1]{Zilin Zhou}
\author[2,4,*]{Zihan Gao}
\author[2]{Yuchen Xie}
\author[1,\ensuremath{\ddagger},\S]{Ran Yi}

\affiliation[1]{Shanghai Jiao Tong University}
\affiliation[2]{Meituan}
\affiliation[3]{National University of Singapore}
\affiliation[4]{Tsinghua University}

\metadata{\sffamily\small\makebox[\linewidth][c]{%
  \href{https://orangerangeangenge.github.io/Demo2Skill/}{\templateicon{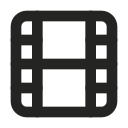}\;\textcolor{metablue}{Homepage}}\qquad
  \href{https://github.com/OrangeRangeAngeNge/Demo2Skill}{\templateicon{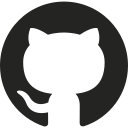}\;\textcolor{metablue}{Code}}}}

\abstract{%
Robo-use agents can accumulate reusable skills, but learning dexterous manipulation from scratch is costly. Egocentric human videos offer interaction priors whose transfer requires interactive scene reconstruction and cross-embodiment motion adaptation. We present \textbf{Demo2Skill}, a real-to-sim-to-real framework that converts demonstrations into executable trajectories and callable skills. Agentic code-based real-to-sim reconstructs an editable scene from video through grounding, Blender refinement, and MuJoCo verification, retaining verified modeling procedures. Agent-driven dexterous trajectory synthesis uses video orchestration, hand trajectory transfer, and contact-aware refinement to adapt demonstrated motion to the robot while preserving task goals. Dexterous skill acquisition adapts verified programs across varied object geometries and uses successful rollouts to train vision-language-action (VLA) policies. Experiments show improved reconstruction and task execution, direct trajectory execution on a physical robot, policy transfer from simulation, and skill reuse by a robo-use agent.}

\begin{document}

\maketitle
\begingroup
  \renewcommand{\thefootnote}{\fnsymbol{footnote}}
  \footnotetext[2]{Equal contribution.}
  \footnotetext[3]{Project lead.}
  \footnotetext[4]{Corresponding author.}
  \footnotetext[1]{Work done during internships at Meituan.}
\endgroup
{
  \vspace{0.15cm}

  \centering
  \begin{minipage}{\textwidth}
    \centering
        \includegraphics[width=\linewidth]{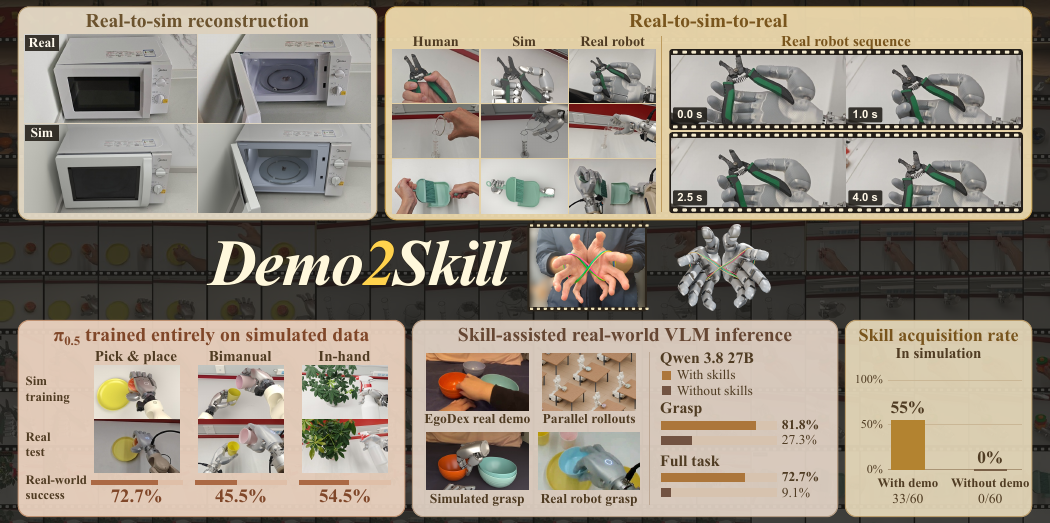}\par
    \captionof{figure}{Overview of Demo2Skill. From an egocentric demonstration, an agent reconstructs an interactive simulation scene and transfers the observed hand--object interaction to dexterous robot trajectories. These trajectories can be executed on a physical robot and expanded into simulation data for training policies that transfer to real-world tasks. The agent also distills successful interactions into reusable skills, which a vision-language agent can invoke for new objects and tasks.}
    \label{fig:demo2skill-teaser}
  \end{minipage}\par
}

\section{Introduction}
\label{sec:introduction}

Powered by large language models (LLMs), robo-use agents have begun to accomplish robotic manipulation by observing environments, generating control programs, and leveraging execution feedback~\cite{liang2023code}\cite{lu2026aspire}\cite{fu2026cap}\cite{huang2026scaling}\cite{xiao2026enpire}. In digital workflows such as coding and writing, agents accumulate capabilities by summarizing past experience into reusable skills rather than solving every problem from scratch~\cite{anthropic2025agentskills}. Bringing this skill-reuse paradigm to physical manipulation is critical for handling diverse everyday tasks, yet physical motions cannot be directly formulated as reusable skills. Egocentric human videos~\cite{hoque2026egodex}\cite{grauman2022ego4d}\cite{hao2026egosim}\cite{elmoghany2026tooldex} offer a rich source of such priors by capturing contact choices, object motion, and bimanual interaction in the wild. Translating these demonstrations into executable, callable skills thus provides a vital pathway for enhancing robotic manipulation capabilities.

Existing approaches for transferring human manipulation experience to robots commonly follow a real-to-sim-to-real pipeline. Video-based methods combine hand--object reconstruction with physical action refinement~\cite{liu2026egoengine}\cite{paliwal2026doasido}\cite{mao2026omnihoi}, but their applicability is constrained by strict reconstruction assumptions and narrow scene coverage. Approaches focusing on morphology adaptation, contact guidance, or multi-reference retargeting \cite{sadjadpour2026morphometric}\cite{zhu2026chord}\cite{lee2026flashdexretarget}\cite{han2026video2sim2real} presuppose available object assets and 3D interaction references, while relation-guided policy learning \cite{lee2026dex} similarly relies on prepared simulation assets and a predefined relation vocabulary. Across these methods, transfer remains bottlenecked by two coupled challenges: recovering physically valid interactive scenes from incomplete visual evidence, and adapting noisy demonstration references across embodiments when direct tracking fails. Motivated by recent advances in agent-driven real-to-sim-to-real methods~\cite{li2026agentic}\cite{wang2026reconstruct}\cite{hu2026simex}, we ask: \textit{Can we construct physically valid scenes from demonstration videos and synthesize executable, reusable manipulation skills under task constraints?}

Our key insight is to represent scenes with executable code and robot motion with parameterized control programs, allowing both to be iteratively refined using demonstration evidence and simulation feedback. In scene construction, code exposes object geometry and physics as editable variables, so visual or physical discrepancies can be resolved through direct program edits. For action synthesis, the agent directly adjusts wrist poses and finger joint targets, then evaluates the resulting motion in simulation. This process adapts execution to the robot's morphology while preserving the task and contact structure of the demonstration. The agent saves verified modeling procedures in a physical modeling skill library and turns verified control programs into callable manipulation skills.

We introduce \textbf{Demo2Skill}, a real-to-sim-to-real framework that converts egocentric human demonstrations into executable robot control programs and reusable skills. Agentic code-based real-to-sim uses grounding, Blender refinement, and MuJoCo verification to reconstruct the scene as parameterized code and iteratively correct its geometry and joints. Agent-driven dexterous trajectory synthesis uses video orchestration, hand trajectory transfer, and contact-aware refinement to generate robot trajectories that satisfy task and contact constraints. Dexterous skill acquisition distills verified control programs into callable skills for robo-use agents and uses randomization and rollout across robot embodiments to generate VLA fine-tuning data. Real-robot trials demonstrate both direct trajectory execution and VLA transfer, while equipping a Qwen 27B agent with a generated skill for grasping thin-rimmed bowls increases its success rate from below 10\% to above 70\%.

To summarize, our contributions are:
\begin{itemize}
    \item \textbf{Agentic code-based real-to-sim.} We propose an agentic approach to reconstructing executable scenes from egocentric demonstrations, iteratively revising scene code using visual evidence and physical feedback to improve geometric and physical consistency.
    \item \textbf{Agent-driven dexterous trajectory synthesis.} We introduce an agent-driven approach that adapts human hand motions into robot-executable trajectories, refining contacts and motions in simulation while preserving the demonstrated task goals.
    \item \textbf{Dexterous skill acquisition.} We distill verified control programs into callable skills and use parameterized scenes to generate diverse trajectories. The resulting trajectories can be used both to train VLA models and to distill robo-use skills that agents can invoke to perform tasks. We validate skill reuse and transfer on real robots.
\end{itemize}

\section{Related Work}
\label{sec:related-work}

\noindent\textbf{Real-to-sim reconstruction for robot manipulation.}
Real-to-sim methods reconstruct real-world environments as physics-based simulations for robot learning and evaluation~\cite{torne2024RialTo}\cite{han2026Re3Sim}\cite{zhang2026robosnap}\cite{ranawaka2026simfoundry}. Recent agentic pipelines coordinate perception, reconstruction, and calibration tools with simulation-based validation, using recorded robot trajectories to detect interaction failures and refine executable scenes~\cite{chen2026agenticReal2Sim}\cite{xu2026worldmirror}. As coding agents become more capable, some methods explore code as an explicit representation for world reconstruction and generation~\cite{shen20264dcodebench}\cite{liang2026visphyworld}\cite{yang2026marc}\cite{yang2026code}\cite{xiao2026beyond}. Besides, Video2World~\cite{tang2026video2world} benchmarks coding agents that construct interactive scenes and robot behaviors from embodied videos. Unlike these methods, Demo2Skill recovers the initial scene as executable code from an entire egocentric video with hand occlusions and object motion.

\noindent\textbf{From human video to robot dexterity.}
Dexterous policies are trained from simulation-based reinforcement learning~\cite{openai2018dexterous}, teleoperation and wearable motion capture~\cite{ding2024bunny}\cite{wang2024dexcap}, and demonstrations synthesized from a few source episodes~\cite{jiang2025dexmimicgen}. Human video is a cheaper source: DexMV and VideoDex use it to guide imitation learning~\cite{qin2022dexmv}\cite{shaw2023videodex}, and ManipTrans, SPIDER, EgoEngine, Video2Sim2Real, and Do as I Do retarget human hand motion into feasible robot trajectories through residual policies or physics-based optimization~\cite{li2025maniptrans}\cite{pan2025spider}\cite{liu2026egoengine}\cite{han2026video2sim2real}\cite{paliwal2026doasido}. These pipelines treat the demonstration as a reference trajectory, so the result is bound to the demonstrated scene and embodiment, and a failed optimization does not say which contact or hand was at fault. Demo2Skill instead has a coding agent write a program that satisfies each operation under physics, yielding an editable program with per-operation verification and a replayable rollout for every accepted trial.

\noindent\textbf{Coding agents for robotics.}
Language models translate natural-language instructions into robot behaviors by generating executable code that composes perception, planning, and control APIs~\cite{liang2023code}\cite{singh2023progprompt}\cite{liu2025hycodepolicy}. Complementary approaches provide geometric and semantic grounding for robotic manipulation, from constructing spatial objectives for low-level controllers~\cite{huang2023voxposer} to automated operation primitive extraction and semantic anchoring~\cite{zheng2026pasg}. Building on these programmatic interfaces, recent coding agents close the loop by executing candidate programs, inspecting execution traces and multimodal observations, and revising code in response to failures~\cite{lu2026aspire}\cite{liu2025hycodepolicy}. Repeated interaction further enables agents to retain successful routines as reusable skills or improve policies and learning procedures through online exploration and real-world experimentation~\cite{huang2026scaling}\cite{xiao2026enpire}. Simulation provides a complementary setting for this process, allowing agents to test and refine robot programs through repeated practice before transferring the resulting skills to physical robots~\cite{hu2026simex}\cite{wang2026reconstruct}\cite{li2026agentic}. In contrast, Demo2Skill focuses on acquiring reusable dexterous skills from egocentric human videos, bridging the embodiment gap between human and robot hands through simulation-based refinement of demonstration-derived trajectories.

\section{Method}
\label{sec:method}

\begin{figure*}[t]
  \centering
   \includegraphics[width=1\linewidth]{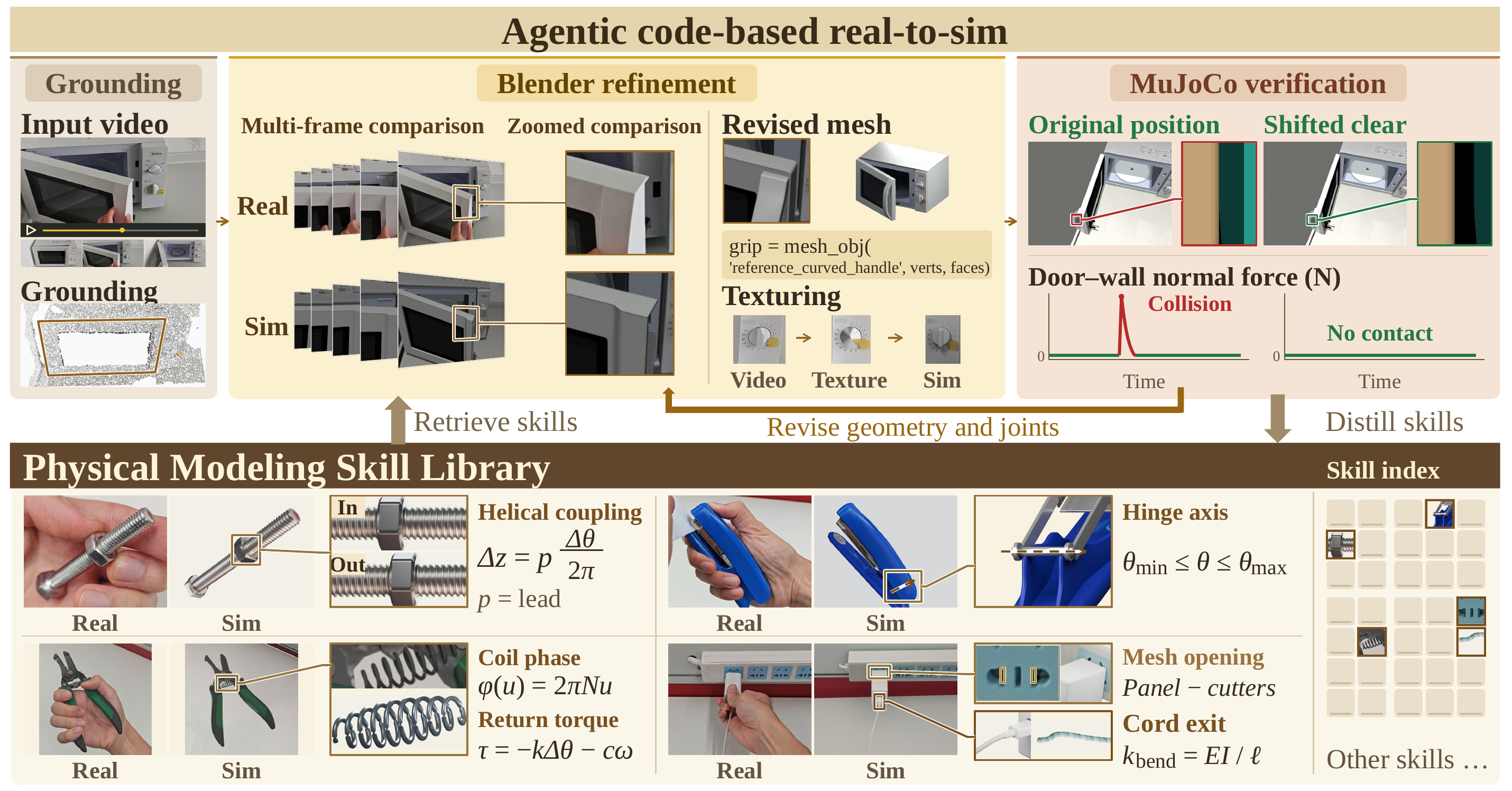}

   \caption{Agentic code-based real-to-sim. Grounding initializes an editable simulation scene from the egocentric video. The agent jointly compares rendered and observed views, enlarges interaction regions, and revises object meshes and textures in Blender. MuJoCo interaction tests expose collisions and incorrect motion, guiding further corrections to geometry and joints. Verified procedures for modeling interactive mechanisms are retained in a physical modeling skill library for subsequent scenes.}
   \label{fig:agentic-real2sim}
\end{figure*}

Demo2Skill comprises three connected modules: agentic code-based real-to-sim, agent-driven dexterous trajectory synthesis, and dexterous skill acquisition. The agent first builds an interactive simulation scene from the demonstration, then generates and verifies robot hand motion in that scene, and finally turns successful interactions into reusable skills.

\subsection{Agentic code-based real-to-sim}

Dexterous manipulation requires a reconstruction with the correct metric scale, enough clearance for the fingers, and moving parts that respond to contact. We use the egocentric demonstration to recover the scene at the start of the task. The agent represents the scene as executable code, so it can change an object's placement, edit a mesh, or revise a joint and then replay the same interaction (Figure~\ref{fig:agentic-real2sim}).

\paragraph{Grounding, Blender refinement, and MuJoCo verification.}
For grounding, we use VGGT~\cite{wang2025vggt}, MoGe3~\cite{kong2026moge3}, and SAM3D~\cite{chen2026sam3d}. Their complementary geometric estimates provide an initial reference for camera and scene layout, object shape, scale, and relative placement. The agent converts this reference into an editable, executable scene whose meshes, textures, object poses, and articulated parts can be revised independently.

During Blender refinement, multi-frame comparison matches rendered views to demonstration frames that reveal different sides and states of the same object. It enlarges the regions around the hand--object interaction to inspect fine geometry, contact clearance, and surface appearance. Discrepancies across the views guide targeted mesh and texture edits: the mesh must explain the visible contours and the space available for the fingers, while the texture must remain consistent with the observed appearance. Comparing several views also helps separate a local shape error from an error in the object's overall pose or the camera alignment.

For MuJoCo verification, the agent replays the demonstrated interaction in the reconstructed scene. Rendering agreement alone cannot establish whether the hand can reach a valid grasp pose or whether a moving part responds as observed. The simulation exposes collisions, insufficient clearance, implausible motion, and incorrect joint placement or constraints. The agent uses these failures to revise geometry, attachments, and joint definitions, then repeats the visual and simulation checks. This feedback loop produces a scene that matches the video while supporting the task's actual interaction.

\paragraph{Physical modeling skill library.}
Visual reconstruction provides an object's appearance and placement, and standard physics checks can reveal issues such as collisions. Completing a manipulation task also depends on finer physical mechanisms. When squeezing spring-loaded pliers, the spring's restoring action determines how the hand must sustain its force. When driving a screw, rotation must advance it along its axis. An agent that builds a scene from appearance and merely checks whether its parts can move may miss these dynamic relations, even though they determine whether the task can be completed.

We therefore maintain a physical modeling skill library organized around interactive mechanisms. The agent uses the demonstrated task and object motion to retrieve relevant skills and construct an executable scene model. It then attempts the task in simulation and uses the interaction outcome to revise connections, motion constraints, and physical responses. When a new task calls for a mechanism that the library does not yet cover, the agent distills the verified modeling procedure into a new skill for later scenes. Scene reconstruction, task execution checks, and skill accumulation together form our real-to-sim pipeline.

\begin{figure*}[t]
  \centering
  \includegraphics[width=1\linewidth]{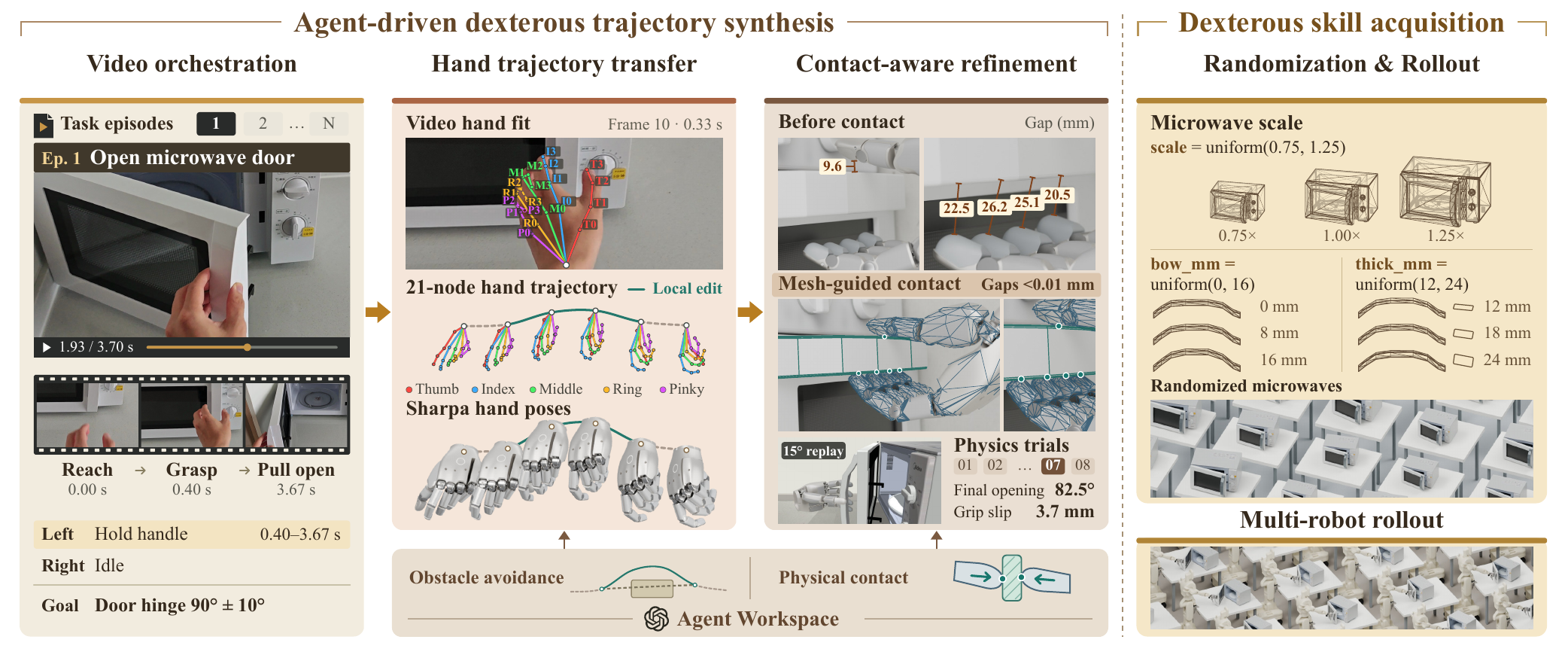}
  \caption{Agent-driven dexterous trajectory synthesis and dexterous skill acquisition. The agent organizes the demonstration into ordered operations with hand roles and contact goals, transfers the observed wrist and finger motion to a robot hand, and refines the grasp using object geometry and physics feedback. It then turns verified operations into a callable skill and tests the skill through rollouts across varied object geometries and robot executions. Successful rollouts provide trajectories for downstream policy training.}
  \label{fig:hand-pipeline}
\end{figure*}

\subsection{Agent-driven dexterous trajectory synthesis}

Given the reconstructed scene, we generate wrist poses and finger joint targets for each robot hand involved in the task. The demonstration provides a reference for how the human hand approaches the object, establishes contact, and moves with it. The reconstructed scene supplies the geometry and dynamics needed to test whether the robot can reproduce the intended change in object state. The first three columns of Figure~\ref{fig:hand-pipeline} illustrate this process.

\paragraph{Video orchestration.}
The agent decomposes the demonstration into operations with explicit temporal order, hand roles, manipulated parts, and intended outcomes. Each operation records how it begins, which contacts it must establish or preserve, and what object state should follow. This representation connects actions that would otherwise be treated as isolated motion segments. For example, reaching a handle prepares the grasp, and the grasp must persist while the hand moves the articulated object. When both hands participate, the same description specifies which hand manipulates the object and which hand stabilizes it.

The operation sequence also determines where trajectory changes are permitted. During approach, the wrist may take a different collision-free route if it reaches a usable grasp pose. At grasp onset, the relative placement of fingers and object becomes critical. During manipulation, the hand must follow the motion induced by the object's mechanism while maintaining contact. By retaining these relationships across operation boundaries, the agent can refine an approach without losing the contact needed by the subsequent action.

\paragraph{Hand trajectory transfer.}
We fit wrist and finger landmarks to the video. The wrist path captures the approach and global hand motion, while the finger landmarks describe how the digits spread and close around the object. Together, they initialize robot wrist poses and joint targets in the reconstructed scene.

Human and robot hands differ in link lengths, joint limits, and reachable fingertip arrangements, so matching the observed wrist path may miss the intended contacts. A kinematic preview compares robot fingertips with the object mesh. The agent revises the wrist approach when the whole hand is misplaced, or adjusts individual fingers when the palm is well positioned but a fingertip misses its surface. This preserves useful demonstrated motion while adapting the grasp to the chosen robot hand, including hands with different degrees of freedom.

\paragraph{Contact-aware refinement.}
At grasp onset, the agent checks the gaps between the robot fingertips and the relevant object surfaces. It uses the object mesh to refine the fingers toward the intended contacts while respecting the robot's kinematics and avoiding undesirable interpenetration. This geometric check provides a grasp candidate, but successful manipulation also requires the contact to remain useful as the object moves.

The agent therefore replays the operation in simulation and examines both the evolving contacts and the resulting object state. For articulated objects, the wrist must follow the motion allowed by the mechanism; for movable objects, the grasp must transmit the required forces without slipping or losing the object. Short replays help identify when contact begins to fail. The agent can then revise the wrist path, finger targets, or timing and replay the operation against its intended outcome.

Task progress and contact stability provide complementary feedback. The achieved object state shows whether the operation advances its goal, while contact behavior reveals how reliably the hand produces that change. The agent evaluates them together, including at the boundaries between operations. In particular, it tests a manipulation from the grasp actually produced by the preceding reach rather than replacing it with a separately prepared pose. The final hand and object states then initialize the next operation. In this way, the generated trajectory forms a continuous, physically executable interaction rather than a collection of individually plausible poses.

\subsection{Dexterous skill acquisition}

The agent turns verified operations into a callable skill that preserves their order, contact requirements, and task success condition. The demonstrated trajectory provides an initial execution, while the skill records the procedure needed to reproduce the interaction on another object instance. It specifies the target part, how the hand establishes and maintains contact, and the object state that marks completion. Wrist targets, grasp configurations, and timing remain adjustable to the new geometry.

During randomization and rollout, the agent varies the reconstructed scene and runs the control program across the resulting object instances to determine where the skill applies (Figure~\ref{fig:hand-pipeline}, right). Variations in overall shape, size, and the geometry of the manipulated part change the reachable approach, finger clearance, and contact locations. The agent inspects failures such as blocked approaches, unstable grasps, or incomplete object motion, revises the relevant parts of the program, and tests it again. It retains the operation structure while adapting the motions needed to satisfy the same task goal.

Each rollout evaluates the complete sequence from approach through task completion. A change that improves an initial grasp may affect contact later in the operation, so the agent checks the outcome across varied scenes rather than judging a single pose in isolation. The resulting skill is a reusable control procedure; each successful rollout also yields a scene-specific trajectory paired with its simulated environment. These trajectories provide diverse training data, while the callable skill can be retrieved and executed by another agent when a related interaction is required.


\section{Experiments}
\label{sec:experiments}

\subsection{Experimental Setup}

\noindent\begin{minipage}[t]{0.60\linewidth}
\vspace{0pt}

We use MuJoCo and NVIDIA Isaac for simulation. Our physical robot is a Gento Luna V2.0 dual-arm platform equipped with a left Sharpa Wave-L-01 hand, a right Sharpa Wave-R-01 hand, and an Intel RealSense D435 camera (Figure~\ref{fig:robot-platform}).

Data processing and trajectory generation run on two eight-GPU servers with 24\,GB of memory per GPU. Policy training uses eight GPUs with 80\,GB of memory per GPU.

Following the experimental protocols of prior work, we evaluate object reconstruction on DexYCB~\cite{chao2021dexycb}, HOI4D~\cite{liu2022hoi4d}, and OakInk2~\cite{zhan2024oakink2} using 25 videos from each dataset.
\end{minipage}\hfill%
\begin{minipage}[t]{0.37\linewidth}
\vspace{0pt}\centering\captionsetup{type=figure}
\includegraphics[width=0.95\linewidth]{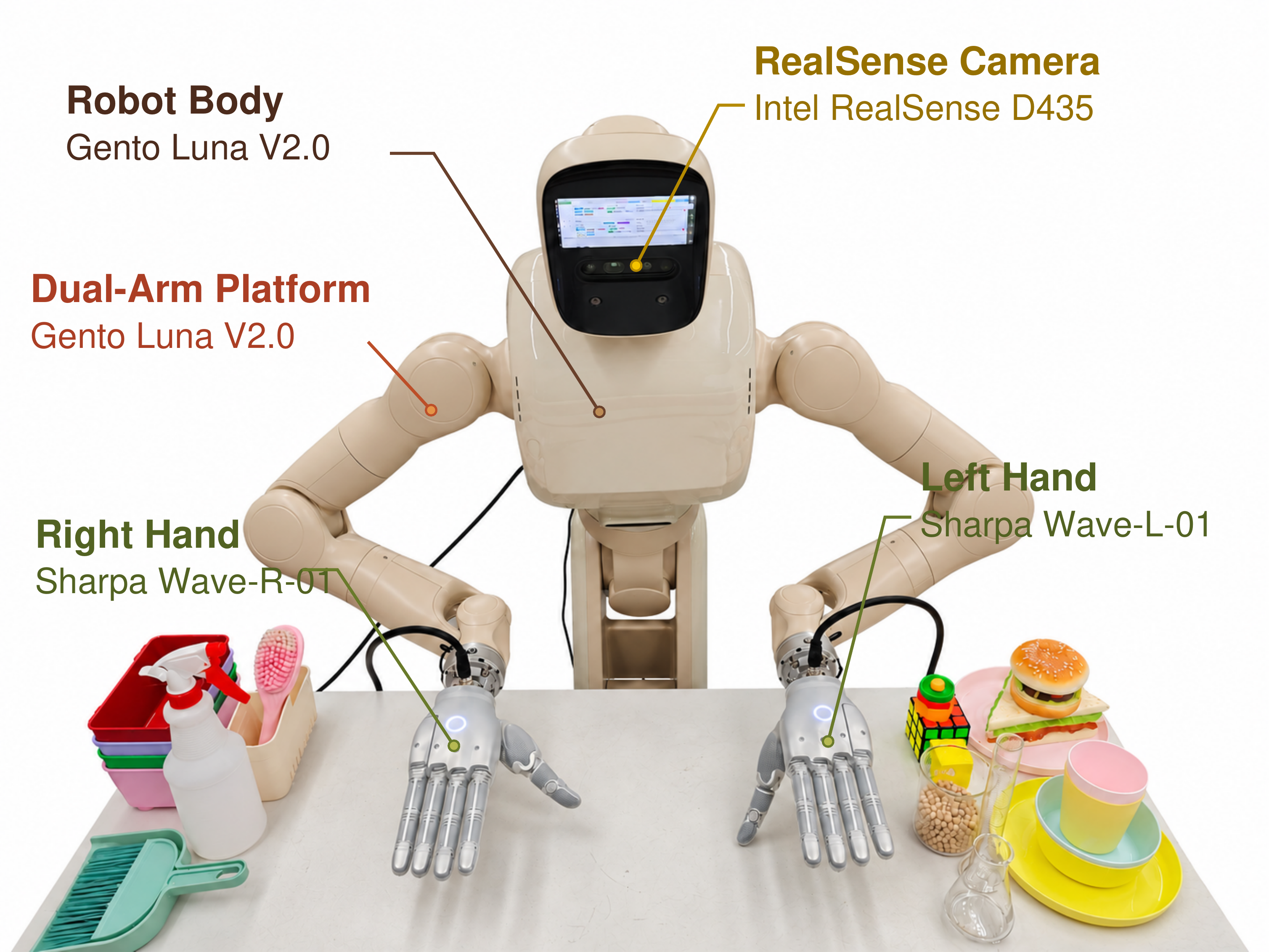}
\caption{Gento Luna V2.0 robot platform.}
\label{fig:robot-platform}
\end{minipage}\par

We evaluate human-to-robot retargeting on 25 OakInk2 videos and end-to-end task execution on 12 EgoDex~\cite{hoque2026egodex} sequences.

\subsection{Human--Object Interaction Reconstruction Results}
We evaluate the reconstruction of objects manipulated by human hands on DexYCB, HOI4D, and OakInk2, using 25 videos from each dataset. Table~\ref{tab:reconstruction} reports F-5, F-10, and Chamfer distance (CD). Our agentic reconstruction approach achieves the highest F-scores and the lowest CD across all three datasets.

\begin{table*}[!htbp]
  \centering
  \caption{Quantitative comparison of reconstruction quality across three datasets. F-scores are reported in
  percentages at thresholds of 0.05 and 0.10, and CD in units of
  $10^{-2}$. Best values are shown in bold.}
  \label{tab:reconstruction}
  \small
  \setlength{\tabcolsep}{4pt}
  \begin{tabular*}{\textwidth}{@{\extracolsep{\fill}}l*{3}{ccc}@{}}
    \hline
    & \multicolumn{3}{c}{\textbf{DexYCB}}
    & \multicolumn{3}{c}{\textbf{HOI4D}}
    & \multicolumn{3}{c}{\textbf{OakInk2}} \\
    \cline{2-4}\cline{5-7}\cline{8-10}
    \textbf{Method}
    & F-5$\uparrow$ & F-10$\uparrow$ & CD$\downarrow$
    & F-5$\uparrow$ & F-10$\uparrow$ & CD$\downarrow$
    & F-5$\uparrow$ & F-10$\uparrow$ & CD$\downarrow$ \\
    \hline
    HO~\citep{hasson2019joint}
    & 31.89 & 57.34 & 4.02
    & 38.96 & 65.50 & 2.72
    & 28.46 & 52.18 & 4.64 \\
    IHOI~\citep{ye2022ihoi}
    & 19.92 & 39.51 & 6.44
    & 16.79 & 31.69 & 6.24
    & 13.22 & 24.29 & 7.75 \\
    HORSE~\citep{prakash2024horse}
    & 20.65 & 38.90 & 6.35
    & 15.60 & 28.96 & 7.33
    & 14.85 & 31.97 & 6.94 \\
    MCC-HO~\citep{wu2026mccho}
    & 33.93 & 58.21 & 3.60
    & 32.77 & 56.95 & 3.13
    & 27.61 & 50.35 & 4.14 \\
    G-HOP~\citep{ye2024ghop}
    & 12.73 & 22.70 & 5.47
    & 41.68 & 68.53 & 1.67
    & 24.69 & 43.87 & 4.94 \\
    FoundationPose~\citep{wen2024foundationpose}
    & 31.86 & 57.26 & 3.57
    & 38.98 & 64.37 & 2.69
    & 30.92 & 53.20 & 4.26 \\
    Any6D~\citep{lee2025any6d}
    & 46.30 & 69.62 & 2.25
    & 47.32 & 71.98 & 1.56
    & 47.31 & 68.19 & 2.91 \\
    Do as I Do~\citep{paliwal2026doasido}
    & 36.45 & 53.97 & 2.42
    & 65.00 & 88.48 & 0.86
    & 49.84 & 61.15 & 2.24 \\
    \hline
    \textbf{Ours}
    & \textbf{57.46} & \textbf{80.68} & \textbf{1.88}
    & \textbf{66.80} & \textbf{89.70} & \textbf{0.78}
    & \textbf{61.99} & \textbf{79.28} & \textbf{2.22} \\
    \hline
  \end{tabular*}
\end{table*}

\subsection{Robot Retargeting Results}
On 25 OakInk2 videos, we evaluate whether the robot completes the demonstrated task and how closely execution matches the reference trajectory. As shown in Table~\ref{tab:tracking}, Demo2Skill succeeds on 23 of 25 videos (92\%), compared with 22 of 25 (88\%) for Do as I Do. Demo2Skill also achieves the lowest position and rotation errors, at 0.036~m and 0.18~rad, respectively, compared with 0.039~m and 0.21~rad for Do as I Do.

\begin{table}[!htbp]
    \centering
    \caption{Human-to-robot retargeting on OakInk2. SR is the task success rate; Pos and Rot are the position and rotation errors relative to the reference trajectory.}
    \label{tab:tracking}
    \begin{tabular}{lrrr}
        \hline
        Method & SR $\uparrow$ & Pos (m)$\downarrow$ & Rot (rad) $\downarrow$\\
        \hline
        Dex-Retargeting~\citep{qin2023anyteleop}  & 4\%  & 0.19 & 1.30 \\
        SPIDER MJWP-act~\citep{pan2025spider}  & 8 \% & 0.14 & 0.53 \\
        SPIDER MJWP~\citep{pan2025spider}      & 24\% & 0.12 & 0.90 \\
        Do as I Do~\citep{paliwal2026doasido}            & 88\% & 0.039  & 0.21 \\
        \hline
        \textbf{Ours}   & \textbf{92\%} & \textbf{0.036} & \textbf{0.18} \\
        \hline
    \end{tabular}
\end{table}

On 12 EgoDex sequences, we evaluate task execution in simulation (Table~\ref{tab:egodex-execution}). Demo2Skill achieves an 83.3\% success rate, compared with 50.0\% for Do as I Do and 41.6\% for CHORD. Generating the robot motions takes 18.6~min with Demo2Skill, 21.1~min with Do as I Do, and 1618.8~min with CHORD. Demo2Skill also records the lowest hand self-penetration rate (H-R, 15.29\%) and the smallest hand self-penetration depth (H-D, 0.542~mm).

\begin{table}[!htbp]
  \centering
  \caption{Task execution on EgoDex. Time is measured for the motion generation stage only. H-R denotes the hand self-penetration rate (\%), and H-D denotes the hand self-penetration depth (mm).}
  \label{tab:egodex-execution}
  \small
  \setlength{\tabcolsep}{3pt}
  \begin{tabular}{lcccc}
    \hline
    \textbf{Method} & SR $\uparrow$ & Gen. time (min) $\downarrow$ 
    & H-R $\downarrow$ & H-D (mm) $\downarrow$\\
    \hline
    CHORD~\citep{zhu2026chord}         & 41.6\% & 1618.8  & 68.99\% & 1.620 \\
    Do as I Do~\citep{paliwal2026doasido}    & 50.0\% & 21.1  & 77.49\% & 2.148 \\
    \textbf{Ours} & \textbf{83.3\%} & \textbf{18.6} & \textbf{15.29\%} & \textbf{0.542} \\
    \hline
  \end{tabular}
\end{table}

\subsection{Real-World Deployment}
\begin{figure*}[!htbp]
  \centering
   \includegraphics[width=1\linewidth]{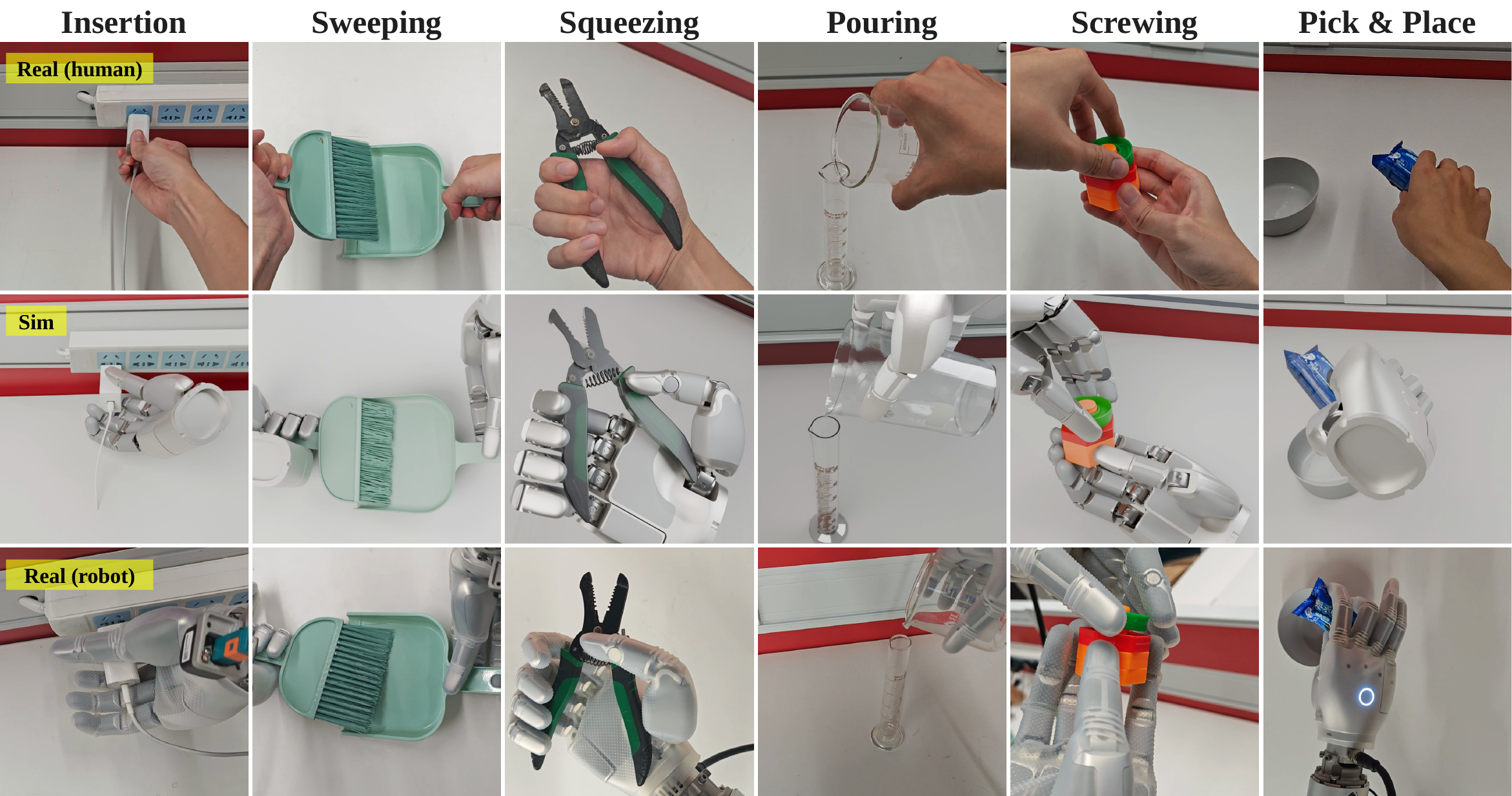}

   \caption{Direct real-to-sim-to-real transfer for grasping, bimanual manipulation, and complex in-hand manipulation.}
   \label{fig:real2sim2real}
\end{figure*}

\paragraph{Direct trajectory execution.}
Our simulated trajectories can be executed directly on a physical robot for grasping, coordinated bimanual manipulation, and complex in-hand manipulation. Figure~\ref{fig:real2sim2real} illustrates this real-to-sim-to-real transfer. We execute the trajectories reconstructed from egocentric videos on the robot.

\paragraph{Policy learning from simulation data.}
We evaluate whether simulation data alone can train a vision-language-action (VLA) policy for real-world manipulation.

\noindent\begin{minipage}[t]{0.52\linewidth}
\vspace{0pt}

We consider three tasks covering single-hand, bimanual, and in-hand manipulation. Each task starts from a short egocentric RGB video recorded with a standard mobile phone. Our pipeline reconstructs high-fidelity simulation scenes as code and synthesizes physically plausible robot trajectories. We vary object poses and parameterized mesh configurations, then run the control program in each modified scene. Each successful rollout contributes a robot trajectory paired with its simulated scene to the training set. The demonstrated task goal stays the same across these variations.

For each task, we use 500 simulated trajectories for supervised fine-tuning of the $\pi_{0.5}$ policy on eight GPUs, then deploy it directly on the physical robot. Figure~\ref{fig:vla-transfer} shows the source demonstrations, randomized simulation scenes, and real-robot executions with their success rates. The policies use only simulated training data.

Across 11 real-robot trials per task, success rates were 72.7\% (8/11) for pick-and-place, 45.5\% (5/11) for bimanual manipulation, and 54.5\% (6/11) for \mbox{in-hand manipulation}.
\end{minipage}\hfill%
\begin{minipage}[t]{0.45\linewidth}
\vspace{0pt}\centering\captionsetup{type=figure}
\includegraphics[width=1\linewidth]{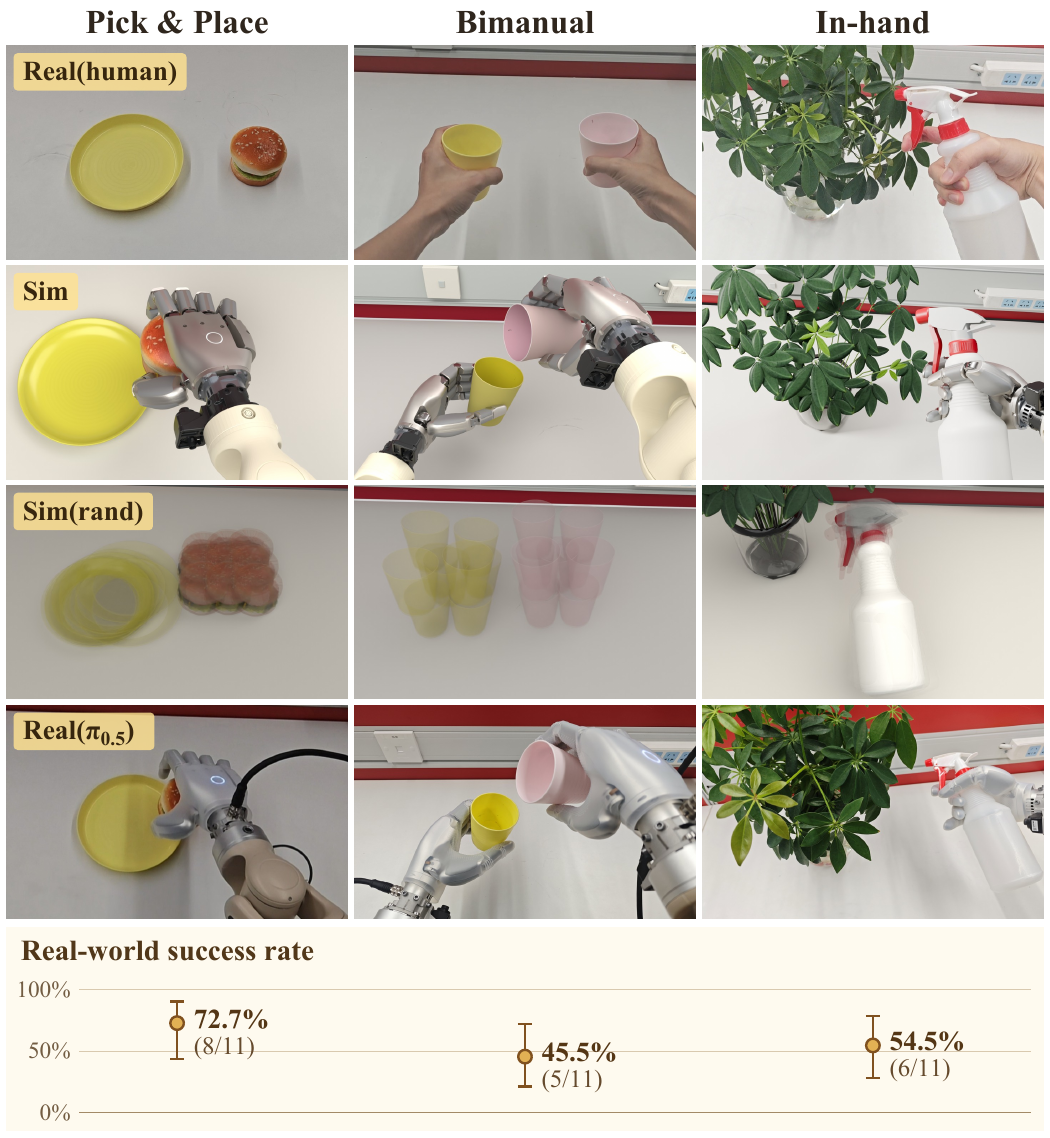}
\caption{Real-world deployment of the $\pi_{0.5}$ policy.}
\label{fig:vla-transfer}
\end{minipage}\par

\subsection{Skill Generalization}

\begin{figure}[!htbp]
  \centering  \includegraphics[width=0.92\linewidth]{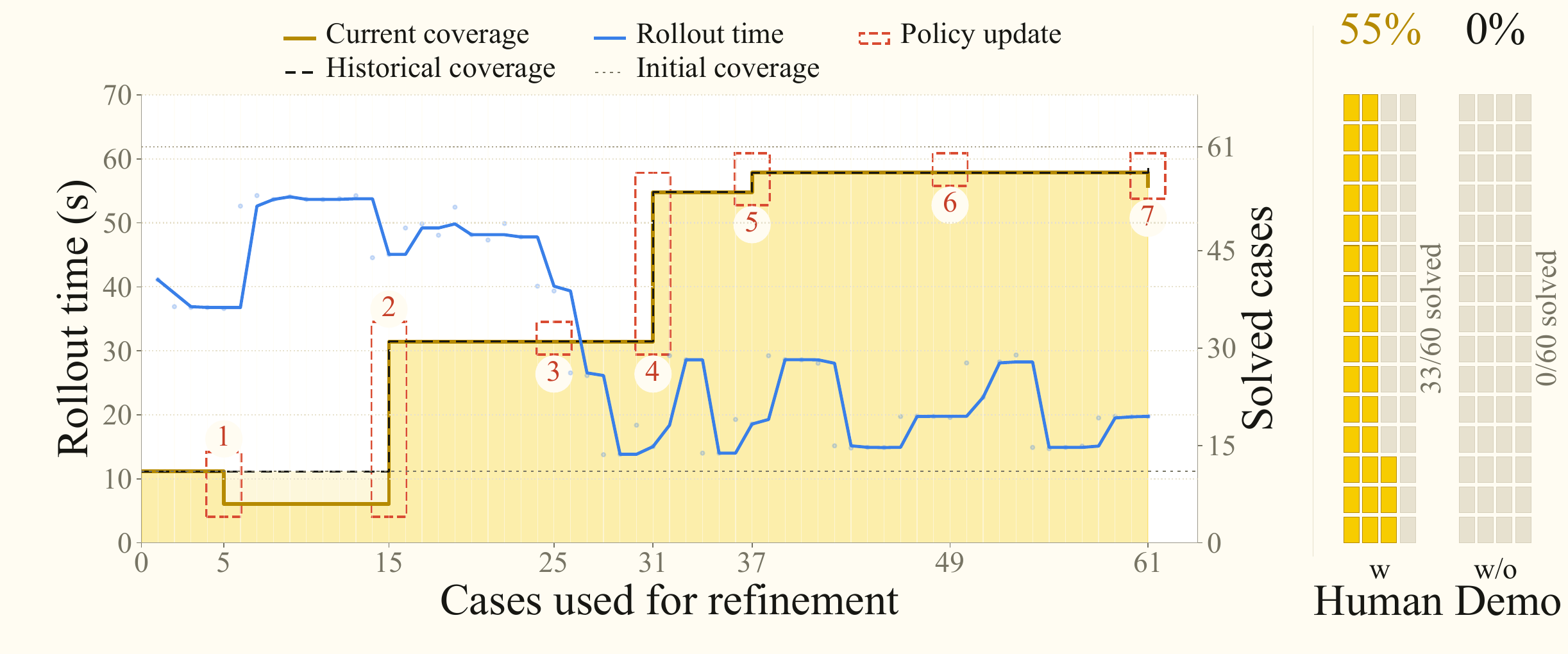}

   \caption{Skill generalization during refinement on randomized simulation cases. Current and historical coverage track the number of cases solved as the skill is updated; the dotted line shows initial coverage, the blue curve shows rollout time, and red markers indicate policy updates.}
   \label{fig:skill-generalization}
\end{figure}

\noindent\begin{minipage}[t]{0.52\linewidth}
\vspace{0pt}

Figure~\ref{fig:skill-generalization} shows how skill coverage grows as the policies are refined in randomized simulation scenes. Together, the four skills initially solved 11 of 61 cases and reached 55 of 61 after refinement. Coverage increased mainly after policy updates. Rollout times varied and were generally lower later. In a separate 60-case evaluation, policies with successful action references achieved 55\% success (33/60), compared with 0\% (0/60) without references under the tested budget.

We also test skill reuse on the physical robot. Figure~\ref{fig:skill-deployment} shows the source egocentric video and skill acquisition in simulation, followed by real-robot trials with and without the specialized skill. We deploy Qwen 27B in a general pick-and-place framework with a generic five-finger pinch-grasping skill. For thin-rimmed bowls, we score the grasp and the full task separately. The full task includes moving the bowl over the target plate. Grasp success rises from 27.3\%~(3/11) with the generic skill to 81.8\%~(9/11) with the specialized skill. Full-task success rises from 9.1\%~(1/11) to 72.7\%~(8/11).
\end{minipage}\hfill%
\begin{minipage}[t]{0.45\linewidth}
\vspace{0pt}\centering\captionsetup{type=figure}
\includegraphics[width=1\linewidth]{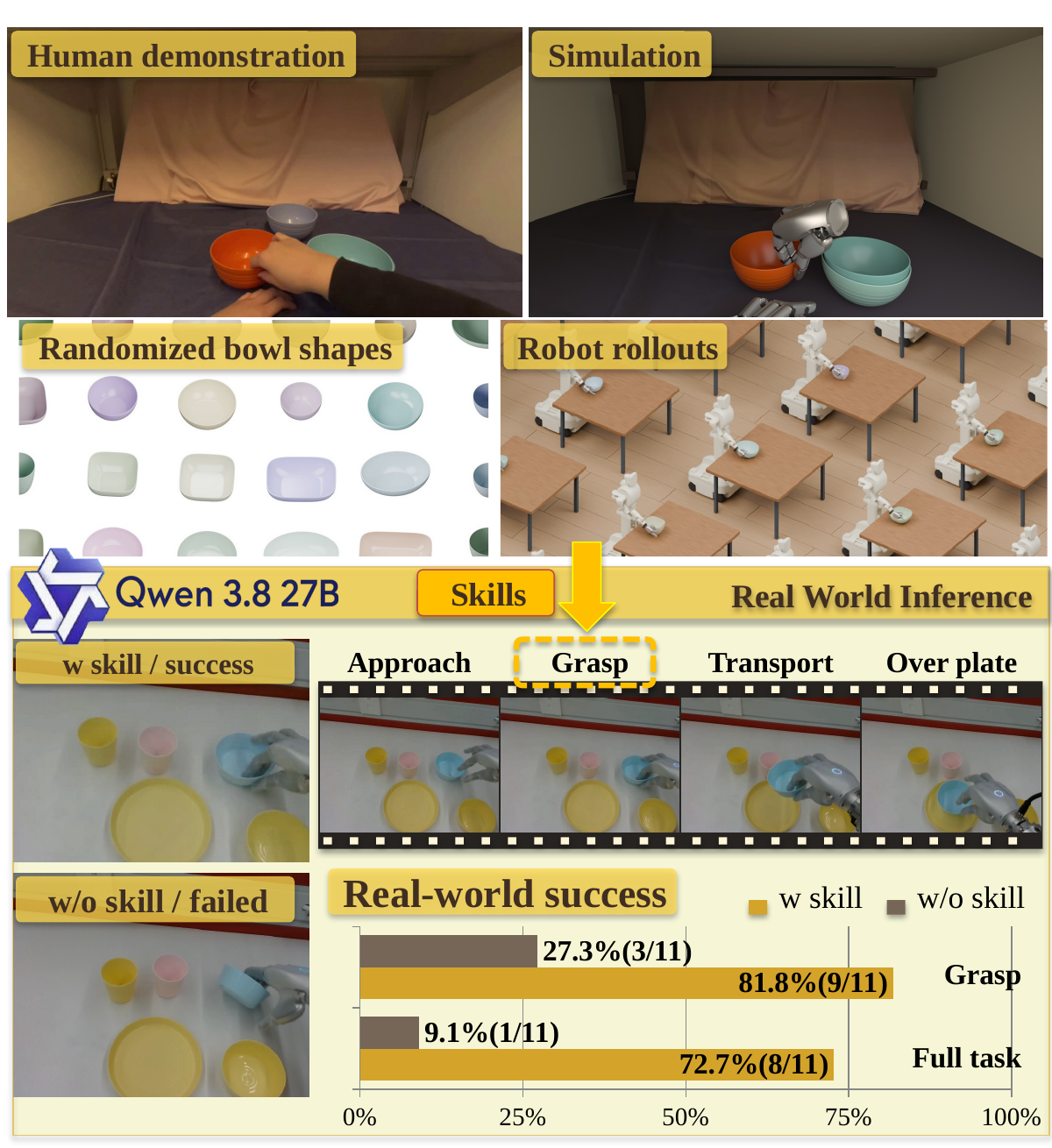}
\caption{Grasping thin-rimmed bowls with Qwen 27B.}
\label{fig:skill-deployment}
\end{minipage}\par

\subsection{Cross-Morphology Evaluation}


We evaluate the same EgoDex tasks with the Wuji, Shadow, and Sharpa Wave hands (Table~\ref{tab:hand-morphology}). Changing the hand morphology leaves the task success rate at 83.3\% for all three hands. Their hand self-penetration rates range from 3.37\% to 15.29\%, well below those of CHORD (68.99\%) and Do as I Do (77.49\%) in Table~\ref{tab:egodex-execution}. The average self-penetration depths are also close across the three hands, ranging from 0.309 to 0.542~mm, and remain below both baselines. These results show that the trajectory generation pipeline transfers across the tested hand morphologies without a decrease in task success or a loss of contact plausibility.

\begin{table}[!htbp]
  \centering
  \caption{Cross-morphology task execution on EgoDex using three dexterous hands with different degrees of freedom (DoF). SR is task success rate; H-R and H-D measure the hand self-penetration rate and mean depth, respectively.}
  \label{tab:hand-morphology}
  \small
  \begin{tabular}{lccc}
    \hline
    \textbf{Hand} & SR $\uparrow$ 
    & H-R $\downarrow$ & H-D (mm) $\downarrow$\\
    \hline
    Wuji Hand (20 DoF)        & 83.3\% & 5.04\% &  0.309 \\
    Shadow Hand (20 DoF)    & 83.3\%   & 3.37\% &  0.318\\
    Sharpa Wave (22 DoF) & 83.3\%  & 15.29\% & 0.542 \\
    \hline
  \end{tabular}
\end{table}

\section{Conclusion}
\label{sec:conclusion}
Demo2Skill turns egocentric demonstrations into dexterous skills that a robot can execute and an agent can reuse. It reconstructs interactive objects at metric scale, uses the observed hand motion to initialize robot trajectories, and refines the grasp in simulation until the robot achieves the demonstrated goal. The verified trajectories transfer to hardware and provide training data for policies that perform pick-and-place, bimanual, and in-hand manipulation on the physical robot. Packaging the same experience as a callable skill enables a vision-language agent to apply the demonstrated manipulation procedure to new task instances. These results show that human demonstrations can yield reusable dexterous robot skills.

\bibliographystyle{unsrtnat}
\bibliography{references}
\end{document}